\documentclass[journal]{IEEEtran}

\usepackage{amsmath,amsfonts}
\usepackage{algorithmic}
\usepackage{array}
\usepackage{textcomp}
\usepackage{stfloats}
\usepackage{url}
\usepackage{verbatim}

\usepackage{graphicx}
\usepackage{subcaption}

\usepackage{cite}
\usepackage{multirow}
\usepackage[pagebackref=true,breaklinks=true,colorlinks,bookmarks=false]{hyperref}
\usepackage{bm}
\usepackage{booktabs}
\usepackage{xcolor,colortbl}
\graphicspath{{fig/}}
\def\eg{\emph{e.g.}}

\def\al{\emph{et al. }}

\begin{document}
\title{Anti-Collapse Loss for Deep Metric Learning Based on Coding Rate Metric}

\author{
Xiruo~Jiang,
Yazhou~Yao,
Xili Dai,
Fumin~Shen,
Liqiang Nie,
Heng-Tao Shen 
	\thanks{X.~Jiang, Y.~Yao are with the School of Computer Science and Engineering, Nanjing University of Science and Technology, Nanjing, China.}
	\thanks{X. Dai is with the Hong Kong University of Science and Technology (Guangzhou), Guangzhou, China.}	
	\thanks{F. Shen and H. Shen are with the School of Computer Science and Engineering, University of Electronic Science and Technology of China, Chengdu, China.}
    \thanks{L. Nie is with the School of Computer Science and Technology, Harbin Institute of Technology (Shenzhen), Shenzhen, China.}
}

\markboth{}%
{Anti-Collapse Loss for Deep Metric Learning}
\maketitle

\begin{abstract}

Deep metric learning (DML) aims to learn a discriminative high-dimensional embedding space for downstream tasks like classification, clustering, and retrieval.
Prior literature predominantly focuses on pair-based and proxy-based methods to maximize inter-class discrepancy and minimize intra-class diversity.
However, these methods tend to suffer from the collapse of the embedding space due to their over-reliance on label information. This leads to sub-optimal feature representation and inferior model performance.
To maintain the structure of embedding space and avoid feature collapse, we propose a novel loss function called Anti-Collapse Loss. Specifically, our proposed loss primarily draws inspiration from the principle of Maximal Coding Rate Reduction.
It promotes the sparseness of feature clusters in the embedding space to prevent collapse by maximizing the average coding rate of sample features or class proxies. 
Moreover, we integrate our proposed loss with pair-based and proxy-based methods, resulting in notable performance improvement.
Comprehensive experiments on benchmark datasets demonstrate that our proposed method outperforms existing state-of-the-art methods. Extensive ablation studies verify the effectiveness of our method in preventing embedding space collapse and promoting generalization performance.
Our code has been available at: \url{https://github.com/NUST-Machine-Intelligence-Laboratory/Anti-Collapse-Loss}. 

\end{abstract}

\begin{figure}[t]
	\centering
	\includegraphics[width=1\linewidth]{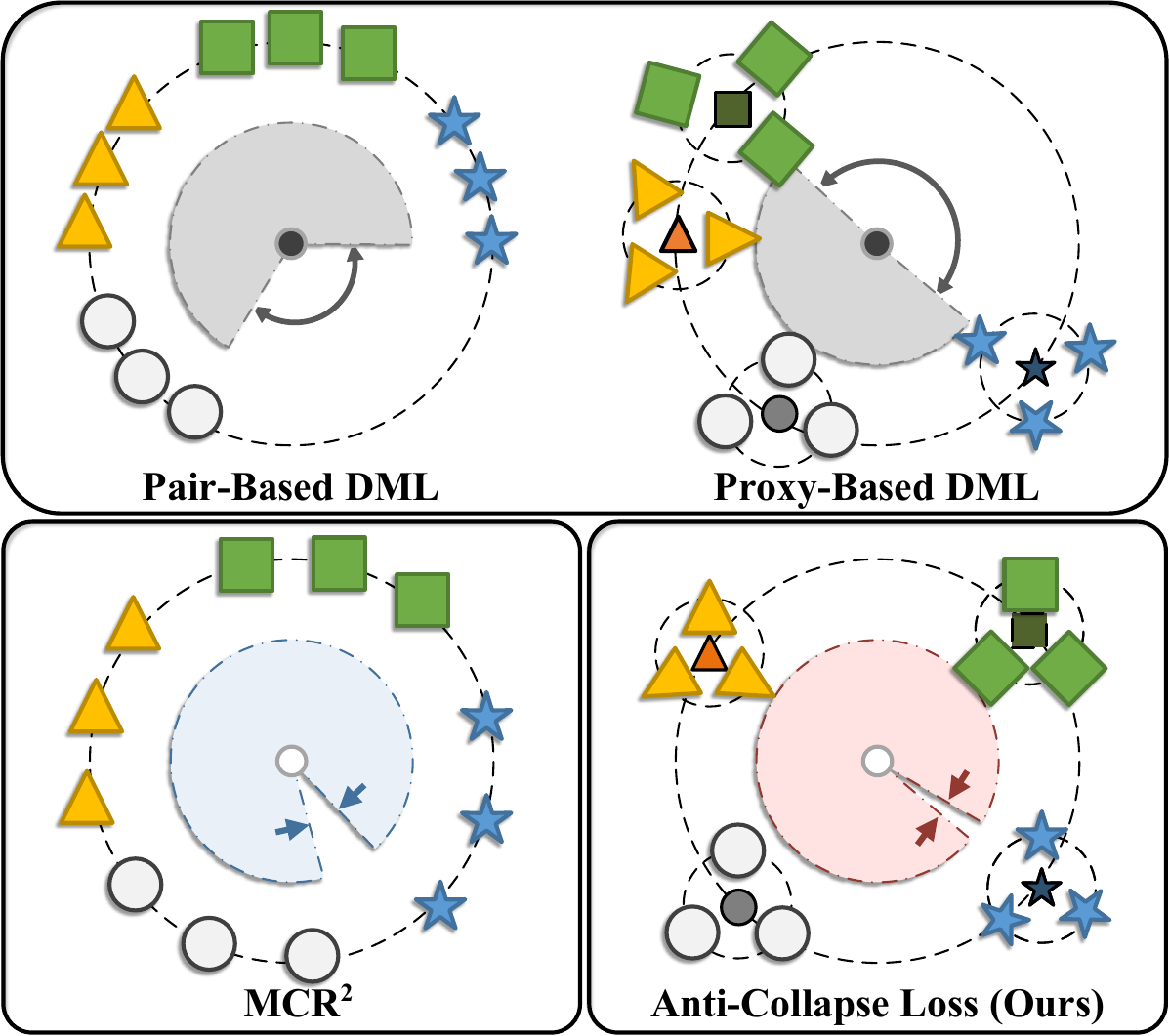}
	\caption{The upper half of the figure illustrates the collapse issue in the embedding space caused by the over-reliance on labels in existing pair-based and proxy-based methods (the sector area represents the coding rate metric of the embedding space). The lower-left subfigure demonstrates the principle of Maximal Coding Rate Reduction (MCR$^2$), which utilizes all data samples to maintain the structure of the embedding space. In contrast, our proposed Anti-Collapse Loss prevents the collapse of embedding space by maximizing the proxy coding rate.}
	\label{motv}
\end{figure}


%
\IEEEpeerreviewmaketitle

\section{Introduction}

\label{sec:introduction}
\IEEEPARstart{L}{earning} compact and generalizable representations has been one of the most critical steps in various machine learning tasks, including image retrieval~\cite{ roth2022non,li2015weakly,wang2023introspective,zheng2021deep,zhao2021towards,yao2020adaptive,jiang2022deep,li2019deep}, face recognition~\cite{Authors21, golwalkar2022masked,huang2021identity}, and image classification~\cite{pei2024cvpr, yao2023automated, yao2021jo,yao2017exploiting,zhang2020web,mao2023attention}, semantic segmentation~\cite{yao2021non, pei2022hierarchical,tang2023holistic,chentaotip2024,10298026,10105896}, few-shot image classification~\cite{li2023deep}.
Focusing on this goal, deep metric learning aims to learn a discriminative embedding space in which distances between semantically similar samples are close while those between dissimilar ones are far apart~\cite{Authors33, Authors34, Authors35,yao2020adaptive,xie2017deep}.

Existing methods can be mainly divided into two groups: pair-based and proxy-based. Pair-based methods seek to learn a better embedding space through relations between sample pairs. For example, \cite{Authors16} proposes to construct an embedding space by employing contrastive loss to enlarge inter-class distances and shrink intra-class distances. Triplet-loss-based methods \cite{Authors21, Authors37, Authors59} propose to adopt sample triplets to assist in metric learning. By employing the triplet of anchor, positive, and negative data points for embedding space optimization, distances between positive pairs are strictly constrained to be smaller than those between negative pairs, thereby improving the generalization performance. However, these methods tend to generate numerous sample pairs or triplets, posing a computation challenge in practical applications.
Another line of research focuses on proxy-based methods for deep metric learning \cite{Authors55, Authors63}. These methods propose to optimize the embedding space by maximizing similarities between sample embeddings and their associated class proxies. Contrary to pair-based methods, proxy-based ones only require class proxy-sample units, whose number is far smaller than sample pairs or triplets, thus resulting in significantly lower computational overhead. Moreover, the elimination of redundant information enhances the generalization performance.
However, similar to pair-based approaches, proxy-based works devote more attention to constructing discriminative embedding space based on label information, leading to reliance on sample annotations.
This issue is prone to causing a collapse in the embedding space during the training process, thereby resulting in sub-optimal model performance.

To this end, we propose a simple yet effective method, dubbed Anti-Collapse Loss, to provide guidance for maintaining the structure of the embedding space. 
Our method is inspired by Information Theory~\cite{cover2006elements} and Maximal Coding Rate Reduction~\cite{yu2020learning}, aiming to reduce reliance on sample labels to avoid embedding space collapse. As shown in Fig.~\ref{motv}, our proposed loss can act as a flexible off-the-shelf module and be seamlessly integrated with existing pair-based and proxy-based methods.

Specifically, we use the Anti-Collapse Loss to handle all the actual samples involved in training, maximizing the average coding rate of sample features within the dataset to prevent the collapse of the embedding space. 
However, this operation also introduces a challenging drawback of Maximal Coding Rate Reduction~\cite{yu2020learning}.
That is, this method requires solving large determinants to estimate covariance, leading to significant computational costs and thereby substantially reducing training efficiency.
To solve the problem of excessive computational consumption, we eliminate the compression term with the highest computational consumption in MCR$^2$~\cite{yu2020learning}. We replace all data samples with class proxies, thus significantly reducing the computation required to solve determinant matrices.
In this way, our Anti-Collapse Loss achieves structural maintenance of the embedding space by maximizing the average coding rate of all class proxies.
Resorting to our proposed loss, learned sample features are distributed reasonably in the embedding space, ensuring the sparseness of feature clusters and thus avoiding space collapse. 
Accordingly, samples of different categories are assured to be maximally disjointed in the embedding space, leading to stronger discriminability. Simultaneously, the embedded space maintained by our proposed Anti-Collapse Loss can effectively avoid overfitting issues, thereby achieving improved generalization performance. Our contributions can be summarized as follows:

(1) We propose a novel loss function, dubbed Anti-Collapse Loss, to prevent the collapse in the process of learning discriminative feature embeddings for conventional DML algorithms. 

(2) Our proposed loss can be integrated with DML methods and address the computational challenge by eliminating the intra-class average coding rate term and replacing data samples with class proxies.

(3) Our proposed loss can advocate the sparseness of feature clusters in embedding space, thereby preventing space collapse and enhancing model generalization performance.

(4) Extensive experiments and ablation studies for DML-based image retrieval tasks on benchmark datasets demonstrate the superiority of our proposed approach. 

The structure of this paper is as follows: In Section~\ref{sec:related_works}, we conduct a detailed and comprehensive review of related studies; Section~\ref{sec:method} provides a detailed introduction to our method, while Section~\ref{sec:experiments} showcases the qualitative and quantitative retrieval results of our method on multiple image retrieval datasets, along with presenting ablation studies. Finally, in Section~\ref{sec:conclusion}, we conclude our work.

\begin{figure*}[t]
	\centering
	\includegraphics[width=\linewidth]{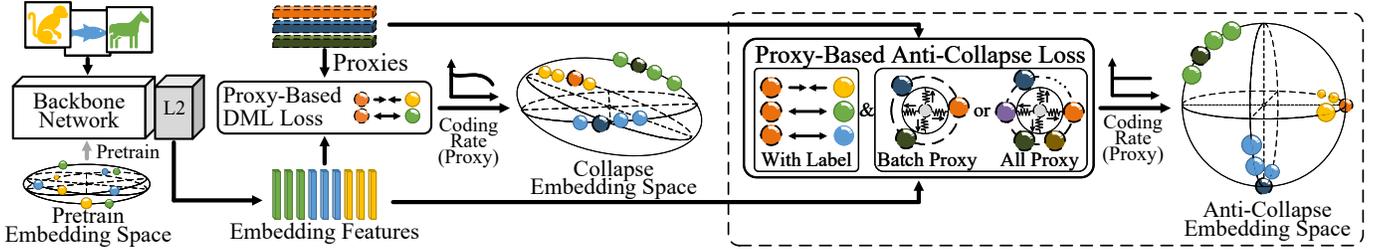}
	\caption{During training, existing proxy-based and pair-based methods require labels to divide samples into positive and negative pairs. These methods rarely utilize the global information of all samples or proxies that have been stripped of label information. The lack of global information leads to the collapse of embedding space during training. To address the collapse issue, our Anti-Collapse Loss maximizes the coding rate to continuously maintain the encoding rate of the proxies during training. Our method fully leverages the characteristic of proxies guiding the sample positions, preventing the collapse of the embedding space while conserving computational resources.}
	\label{fig:pipline}
\end{figure*}

\section{Related Work}
\label{sec:related_works}

\subsection{Manifold Learning}

Manifold learning aims to discover and model low-dimensional manifold structures from high-dimensional data. When manifolds are linear, subspace clustering methods~\cite{ma2007segmentation, elhamifar2013sparse} can be used to cluster the manifolds. There are various methods for handling nonlinear multi-manifold clustering problems, among which the Maximum Code Rate Reduction (MCR$^2$) principle~\cite{yu2020learning} is one of the most representative works in recent years. 
MCR$^2$ estimates the overall compactness of finite samples using the distortion rate concept in information theory. It optimizes the embedding space by leveraging the difference between the overall compactness scale and the sample-averaged coding rate which describes the intra-class compactness of specific categories.

Our proposed method draws inspiration from the MCR$^2$ principle and utilizes the global average coding rate term from this algorithm to optimize the embedding space. Our goal is to optimize the embedding space structure, reduce the influence of class labels on the embedding space, and make our method more generalizable. 
Since our proposed method already has a deep metric learning clustering term, to avoid duplicating the functionality of modules, we remove the compression term from MCR$^2$.
Meanwhile, we also notice the significant computational cost associated with estimating the covariance in MCR$^2$. This cost arises from the need to calculate the determinant of large matrices using all samples.
Our proposed Anti-Collapse Loss can effectively address this issue by leveraging proxies. 

\subsection{Deep Metric Learning}
The purpose of metric learning is to accurately measure the similarity between samples in a high-dimensional data space, thereby creating a distance relationship where samples of the same class are closer while samples of different classes are farther apart. This inter-sample relationship is important for tasks like classification and clustering. Therefore, metric learning methods have practical value in tasks such as image retrieval and classification.

However, early metric learning~\cite{Authors35,mensink2012metric,verma2012learning,tsagkatakis2011online} has some drawbacks. Firstly, it often relies on manually designed features, which may be challenging to capture the advanced representations of data. When dealing with complex and high-dimensional data, manually selecting appropriate features can become difficult. Secondly, some traditional metric learning methods may exhibit lower efficiency when handling large-scale data, especially in scenarios requiring pairwise comparisons, leading to a significant increase in computational complexity. Additionally, some traditional metric learning methods may lack good generalization ability in new domains or tasks because they overly focus on specific data representations and metric methods. 

With the flourishing development of deep learning~\cite{yang2019deep,gu2018recent,lu2015rating,yang2024ijcv,yang2021deconfounded,sun2022pnp}  and the rapid progress of computational acceleration hardware, metric learning is gradually transcending traditional limitations. Deep learning has significantly benefited metric learning by efficiently modeling complex data relationships and enabling end-to-end feature representation learning. Convolutional Neural Networks~\cite{krizhevsky2012imagenet,DL7_he2016deep, DL8_ioffe2015batch} have improved the ability of models to learn hierarchical features, making them better suited to handle intricate data structures. With end-to-end learning, models can automatically learn features without the need for manual feature design, thus significantly increasing their expressive capability.

At the same time, the rapid development of computational acceleration hardware has greatly accelerated the training process on large datasets, thereby eliminating the computational bottlenecks. The parallel computing capabilities of Graphics Processing Units (GPUs) have notably accelerated the training of deep learning models, making it easier for these models to handle complex nonlinear data relationships. 

By introducing deep learning, metric learning can achieve end-to-end learning of data representations, reduce dependence on manually designed features, enhance the model's generalization ability, and better adapt to large-scale datasets.
Deep metric learning aims to learn an embedding space using deep neural networks, where complex high-dimensional data is mapped into low-dimensional features~\cite{seidenschwarz2021learning,fu2021deep,liu2022densely,roth2021simultaneous,kan2022contrastive,ko2021learning,jiang2024dual}. 
Similar samples are closer to each other in this space, while different class features are apart. Most methods in this field can be divided into two groups: pair-based~\cite{qi2020simple} and proxy-based~\cite{Authors55,Authors63,gu2021proxy,zheng2023deep}. Pair-based methods directly optimize the distance or similarity between samples. The Siamese network~\cite{Authors14} is among the early works introducing deep learning into metric learning. Most deep metric learning loss functions are inspired by contrastive~\cite{Authors16} and triplet loss~\cite{Authors21}. MS Loss~\cite{Authors7} combines multiple types of similarities, considering both intra-class and inter-class relationships more comprehensively. On the other hand, proxy-based methods utilize the metric relationship between category proxies and samples for classification and clustering. ProxyNCA~\cite{Authors55} and ProxyAnchor~\cite{Authors63} are representative methods in this category. The advantage of proxy-based methods lies in their computational efficiency and scalability. Compared to pair-based methods that require pair comparisons, the complexity of distance computation is significantly reduced by using proxies as representatives of samples.

To address the collapse issue in existing metric learning methods caused by reliance on label information, our proposed Anti-Collapse Loss optimizes the structure of the embedding space by maximizing the coding rates of all samples or proxies. Additionally, we incorporate the concept of proxies into the coding rate function to enhance the model generalization ability while maintaining a lightweight property for the coding rate function.

\section{The Proposed Method}
\label{sec:method}
Existing DML methods emphasize the construction of embedding spaces. However, they pay less attention to accurately representing the inherent geometric or statistical characteristics of the sample feature distribution within the embedding space.
This leads to a situation where existing methods excessively rely on labelled data without using sufficient knowledge of the underlying data distribution. Such reliance on labelled data negatively impacts model generalization ability.
We design a new loss function based on the coding rate to prevent the collapse of the embedding space. Specifically, our proposed Anti-Collapse Loss significantly enhances the coding rate differences between the entire dataset and clusters of different classes while reducing the explicit dependence of the model on labels. As shown in Fig.~\ref{fig:pipline}, we present the Anti-Collapse Loss's functionality in maximizing sample and proxy coding rates during the training process. Additionally, we demonstrate the space collapse issue caused by existing pair-based and proxy-based deep metric learning algorithms. 

We assume that the backbone network $\mathcal{B}(\mathcal{S},\theta)$ extracts feature $x_{i}^{*}$ from sample $s_{i}\in \mathbb{R}^D$, where $i\in \{1,2,...,N\}$, $\mathcal{S}=\{s_1,s_2,...,s_N\}$ represents the dataset with $\mathcal{S} \in \mathbb{R}^{N \times D}$, $\theta$ denotes the network parameters, and $N$ represents the number of samples. Each sample $s_i$ corresponds to a class label $y_i \in \mathcal{Y}$, where $\mathcal{Y}=\{y_1,y_2,...,y_m\}$. Next, we project the feature $x_{i}^{*}$ onto a unit hypersphere embedding space with normalization. The normalized features are represented as $\mathcal{X}=\{x_1,x_2,...,x_N\}$, where $\mathcal{X} \in \mathbb{R}^{N \times d}$ and $d$ represents the dimensionality of the features.

\subsection{The Rate Distortion Function} \label{The Rate Distortion Function}

The optimal coding methods for independently and identically distributed (i.i.d.) samples with a known probability distribution $p(\mathcal{X})$ have been extensively analyzed and explored in information theory~\cite{cover2006elements}. However, the enrichment and complexity of data in classification tasks reduce the applicability of optimal coding methods based on the probability distribution $p(\mathcal{X})$. For example, most classification tasks based on deep convolutional networks currently rely on features $\mathcal{X}$ from limited training samples with an unknown probability distribution $p(\mathcal{X})$.
Nevertheless, recent works find that certain fundamental concepts used to derive the optimal coding rate can still be applied to estimate the coding rate.
Ma \al proposed the nonasymptotic rate distortion in MCR$^2$~\cite{ma2007segmentation, yu2020learning} to accurately estimate the number of bits required for coding finite samples from class-subspace distributions. 

For our work, in the practical training task of image retrieval, according to nonasymptotic rate distortion, we can obtain the equation for the average Gaussian coding rate of features as follows:
\begin{equation}
	\label{mean bits XTX}
	\begin{aligned}
		R(\mathcal{X}_{bs}, \varepsilon) = & \frac{1}{2} \log \det \left( I + \frac{d}{n_{bs}\varepsilon^2} \mathcal{X}_{bs}^\top \mathcal{X}_{bs} \right), \\ &\mathcal{X}_{bs}^\top \mathcal{X}_{bs} \in \mathbb{R}^{d \times d},
	\end{aligned}
\end{equation}
where $n_{bs}$ represents the batch size, $\varepsilon$ represents precision and $\mathcal{X}_{bs}$ represents the features of a current batch in the iteration. The notation $\log \det$ represents the use of the logarithm of the determinant as a smooth approximation for rank in solving rank minimization problems. It guarantees convergence to a local minimum \cite{fazel2003log}.

Then, according to Eq.(\ref{mean bits XTX}) and the Minimum Description Length criterion~\cite{barron1998minimum}, we can estimate the total number of bits required for features learned by a deep network with a quantity of $N$ and an embedding dimension of $d$, i.e., the rate-distortion coding rate $\Gamma(\mathcal{X}, \varepsilon)$:
\begin{equation}
	\label{eq:overal coding length}
	\begin{aligned}
		\Gamma(\mathcal{X}, \varepsilon) & = (N+d)R(\mathcal{X}, \varepsilon) \\& = \frac{N+d}{2} \log \det \left( I + \frac{d}{N\varepsilon^2} \mathcal{X}^\top \mathcal{X} \right).
	\end{aligned}
\end{equation}
According to the commutative property of the coding length function introduced in~\cite{ma2007segmentation}, the matrices $\mathcal{X}_{bs}\mathcal{X}_{bs}^\top$ and $\mathcal{X}_{bs}^\top\mathcal{X}_{bs}$ share the same nonzero eigenvalues, the average coding length function can also be expressed as follows:
\begin{equation}
	\label{mean bits XXT}
	\begin{aligned}
		R(\mathcal{X}_{bs}, \varepsilon) = & \frac{1}{2} \log \det \left( I + \frac{d}{n_{bs}\varepsilon^2} \mathcal{X}_{bs}\mathcal{X}_{bs}^\top \right), \\ & \mathcal{X}_{bs}\mathcal{X}_{bs}^\top \in \mathbb{R}^{N \times N}.
	\end{aligned}
\end{equation}
$R(\mathcal{X}, \varepsilon)$ can be used to assess the compactness of the embedding space during training. Our work focuses on utilizing this coding rate measure to avoid or mitigate the issue of collapsing volume in the embedding space during training.

By referring to Eq.(\ref{mean bits XXT}), we can observe that when the sample features are normalized, $ \mathcal{X}_{bs}\mathcal{X}_{bs}^\top$ can be represented as $Sim (\mathcal{X}_{bs},\mathcal{X}_{bs})$ (abbreviated as $Sim (\mathcal{X}_{bs})$). The cosine similarity $Sim (\mathcal{X}_{bs})$ is an important metric in metric learning for representing the relationships between samples. Therefore, we can rewrite the equation as follows:
\begin{equation}
	\label{eq:mean bits cos}
	\begin{aligned}
		R_{pair} = & \frac{1}{2} \log \det \left( I + \frac{d}{n_{bs}\varepsilon^2}  Sim(\mathcal{X}_{bs}) \right),\\&O(R_{pair})=  n_{bs}^2,
	\end{aligned}
\end{equation}
where $O(R_{pair})$ represents training complexity. This equation establishes a connection between the optimization problem of information-theoretic coding rate and deep metric learning, allowing us to optimize the embedding space by controlling the coding rate. Based on this attribute, we leverage it to design a new loss function, the equation of which is as follows:
\begin{equation}
	\label{eq:ac_pair}
	\begin{aligned}
		\mathcal{L}^{pair}( \mathcal{X}_{bs},\varepsilon) = -R_{pair}.
	\end{aligned}
\end{equation}
This loss can not only be combined with supervised deep metric learning methods using pairs or sample tuples, but also serve as an independent loss for unsupervised training to accomplish classification or clustering tasks.

\subsection{Proxy-Based Anti-Collapse Loss}

The Eq.(\ref{eq:ac_pair}) can maximize the utilization of sample information to compute the average encoding rate of samples. However, it also need to face one of the drawbacks mentioned earlier in deep metric learning methods based on pairs: too many pairs are involved, resulting in redundant information and high computational complexity. In contrast, proxy-based methods leverage only a small number of class proxies and relationships between samples, which leads to lower training complexity, better generalization, and faster convergence during training process. For example, ProxyNCA~\cite{Authors55} employs only one real sample feature as the anchor sample $x_{a}$ and replaces the traditional positive sample $x^{+}$ and negative sample $x^{-}$ from the triplet structure with positive proxy $p^{+}$ and negative proxy $p^{-}$. The equation is as follows:
\begin{equation}
	\label{eq:pnca}
	\begin{aligned}
		\mathcal{L}_{PNCA} = - \log \left( \frac{{e^{(Sim(x_{a},p^{+}))}}}{{\sum\limits_{p^{-} \in \mathcal{P}^{-}} e^{(Sim(x_{a},p^{-}))}}} \right).
	\end{aligned}
\end{equation}
However, the ProxyNCA has a significant problem in terms of tuple composition. The real sample feature $x_a$ in the triplet ($x_a$, $p^+$, $p^-$) cannot establish a direct relationship between real samples, like methods based on pairs or tuples. This tuple structure provides better robustness when dealing with adversarial perturbations and significantly reduces the amount of information the model can learn from real data. ProxyAnchor~\cite{Authors63} addresses this issue by constructing triplet structures based on proxy points in the opposite way compared to the ProyxNCA. 
The equation for ProxyAnchor (PA) is:
\begin{equation}
	\label{eq:pa}
	\begin{aligned}
		\mathcal{L}_{PA} &= \frac{1}{|\mathcal{P}^+|} \sum_{p \in \mathcal{P}^+} \log \left( 1 + \sum_{x \in \mathcal{X}_{bs}, y_x=y_{p}} e^{-\alpha \cdot [Sim(x,p_a) - \delta]} \right) 
		\\&+ \frac{1}{|\mathcal{P}|} \sum_{p \in \mathcal{P}} \log \left( 1 + \sum_{x \in \mathcal{X}_{bs}, y_x \neq y_{p}} e^{\alpha \cdot [Sim(x,p_a) + \delta]} \right),
	\end{aligned}
\end{equation}
where $\delta$ is a margin and $\alpha$ is a scaling factor. 

Inspired by proxy-based methods, we construct a new proxy-based loss for efficient coding rate calculation, significantly reducing computational overhead. The equation is as follows:
\begin{equation}
	\label{eq:mean bits cos pxy}
	\begin{aligned}
		R_{proxy}  = & \frac{1}{2} \log \det \left( I + \frac{d}{n_{p}\varepsilon^2}  Sim(\mathcal{P}) \right) ,\\O(R_{proxy}(\mathcal{P}_{bc}))&=n_{bc}^2~or ~O(R_{proxy}(\mathcal{P}_{ac}))=n_{ac}^2,
	\end{aligned}
\end{equation}
where $n_p$ represents the number of proxies, $\mathcal{P}_{bc}$ represents the corresponding proxy set of classes contained in the current mini-batch of data samples, and $\mathcal{P}_{ac}$ represents the proxies for all classes. $n_{bc}$ and $n_{ac}$ represent the number of proxies in two different proxy sets respectively. The proxy $\mathcal{P}$ are artificially defined sets of features. Therefore, without establishing a connection with the feature points of the dataset's samples, they cannot directly influence the overall structure of the data. Hence, Eq.(\ref{eq:mean bits cos pxy}) needs to be combined with a new function based on proxy for constructing a deep metric learning loss function. We thus define a Proxy-based Anti-Collapse Loss as follows:
\begin{equation}
	\label{eq:ac_p}
	\begin{aligned}
		\mathcal{L}_{AntiCo}^{proxy}(\mathcal{P}, \mathcal{X},\varepsilon) &= -R_{proxy}(\mathcal{P},\varepsilon) + \nu \mathcal{L}_{proxy}(\mathcal{P},\mathcal{X}),
	\end{aligned}
\end{equation}

where $Anti$-$Collapse$ is abbreviated as $AntiCo$ and $\nu$ represents the weight of the proxy-based loss $\mathcal{L}_{proxy}$ (\eg, $\mathcal{L}_{PNCA}$ in Eq.(\ref{eq:pnca})). According to Eq.(\ref{eq:mean bits cos pxy}), we divide $\mathcal{L}_{AntiCo}^{proxy}$ into $\mathcal{L}_{AntiCo}^{proxy(All-Class)}$ and $\mathcal{L}_{AntiCo}^{proxy(Mini-Batch)}$ based on our selection of proxies. By comparing the training complexities in Eq.(\ref{eq:mean bits cos}) and Eq.(\ref{eq:mean bits cos pxy}), we can observe that our proposed Proxy-based Anti-Collapse Loss greatly reduces the computational complexity of algorithms, especially when using only the number of sample classes in the mini-batch. In addition, the proxies processed by Eq.(\ref{eq:ac_p}) can maintain a huge inter-class distance gap in the embedding space, which allows the proxy to have better guiding ability for sample classification.

Compared to the rate reduction function in MCR$^2$ and existing pair-based deep metric learning methods, the Anti-Collapse Loss demonstrates significant superiority. It leverages the characteristics of proxies, significantly reducing computational complexity by using sample-proxy and proxy-proxy pairs as training units. Additionally, the Anti-Collapse Loss enhances the guidance capabilities of proxies in existing proxy-based methods by boosting the average encoding rate. These proxies effectively maintain the structure of the embedding space, enhancing discriminability among samples from different classes and consequently improving image retrieval performance.

\section{Experiments}
\label{sec:experiments}

\subsection{Experimental Settings} 

\textbf{Datasets Settings:} We select three commonly used datasets for image retrieval experiments in metric learning: CUB200-2011 (CUB200)~\cite{Authors46}, Cars196~\cite{Authors47}, and Stanford Online Products (SOP)~\cite{Authors49}. For CUB200 dataset, we use the first 100 classes, which include 5,864 images for training, while the remaining 100 classes, with 5,924 images, are reserved for testing. We divide the Cars196 dataset into two parts: 8,054 images from the first 98 classes are used for training, and 8,131 images from the remaining 98 classes are used for testing. For Stanford Online Products dataset, we allot 59,551 images derived from 11,318 classes for training, whereas the remaining 60,502 images from 11,316 classes are utilized for testing.

\textbf{Image Preprocessing:} To ensure fairness comparison, we preprocess the training set images based on the experimental settings of the majority of existing deep metric learning research (\eg,~\cite{Authors7, Authors63}). Initially, all input images are resized to dimensions of 256$\times$256 and horizontally flipped. Subsequently, these images are randomly cropped to a size of 224$\times$224. Similarly, following previous works, the test images are processed using a central cropping operation. 

\textbf{Proxies Settings:} Based on existing proxy-based deep metric learning methods~\cite{Authors55, Authors63}, we allocate one proxy per semantic class in the dataset, and initialize the proxies using a normal distribution.

Hyperparameter Settings: In the experiments, the hyperparameters in Eq.(\ref{eq:ac_p}) are set to $\varepsilon=0.5$, $\nu\in[0.001,0.1]$. The hyperparameters in the $\mathcal{L}_{proxy}$ are set to $\delta=0.1$, $\alpha=32$, by following~\cite{Authors63}.

Hardware Configuration: All experiments are conducted on two 24GB NVIDIA GeForce RTX 3090 GPUs.

\textbf{Backbone and Parameters:} ResNet50 (R50)~\cite{DL7_he2016deep} and BN-Inception (IBN)~\cite{DL8_ioffe2015batch} have been commonly used as backbone networks in deep metric learning. This work also selects these two networks as the backbone. Both networks are loaded with pretrained model parameters trained on ImageNet~\cite{russakovsky2015imagenet}. We can obtain different embedding dimensions for sample features $\mathcal{X}$ by varying the size of the last fully connected layer in the backbone network. We use Adam as the optimizer for the image retrieval experiments to train the backbone network, with a learning rate of $10^{-5}$. We employ a large learning rate multiplication strategy for training the proxies. Except for the experiments comparing image retrieval performance under different batch sizes, the batch size for each experiment is set to 90.

\textbf{Evaluation Criteria:} To measure the performance of Anti-Collapse Loss in image retrieval tasks, we evaluate the method using Recall@$K$(\%) and Normalized Mutual Information (NMI) score. Recall@$K$(\%) refers to the percentage of images that retrieve at least one correctly matched sample from the top $K$ nearest neighbors, while the NMI score is obtained by computing the ratio of mutual information to the average entropy between clustering results and ground truth labels.

\begin{table*}[t]
	\renewcommand{\arraystretch}{1.1}
	\setlength\tabcolsep{7pt}
	\footnotesize
	\centering
	\resizebox{1\textwidth}{!}{
		\begin{tabular}{l | c | c | c | c | c | c | c }
			\bottomrule
			\multicolumn{2}{c|}{Datasets} & \multicolumn{3}{c|}{\shortstack{\\[0.1ex]CUB200}} & \multicolumn{3}{c}{Cars196} \\
			\toprule
			\bottomrule
			\textsc{Methods} & Arch/Dim. & R@1 & R@2 & NMI & R@1 & R@2 & NMI\\
			\toprule
			\bottomrule
			ProxyNCA~\cite{Authors55} $_\textrm{ICCV2017}$        & IBN/64 & 49.2 & 61.9 & - & 73.2 & 82.4 & -    \\
			MS~\cite{Authors7}$_\textrm{CVPR2019}$             & IBN/64 & 57.4 & 69.8 & - & 77.3 & 85.3 & -    \\
			SoftTriple~\cite{qian2019softtriple} $_\textrm{ICCV2019}$  & IBN/64 & 60.1 & 71.9 & - & 78.6 & 86.6 & -   \\ 
			ProxyAnchor ~\cite{Authors63}$_\textrm{CVPR2020}$ & IBN/64 & 61.7 & 73.0 & - & 78.8 & 87.0 & -     \\ 
			\hline
			\rowcolor{gray!25}Anti-Collapse               & IBN/64 & 62.4 & 74.1 (+1.1) & 67.1 & 80.1 & 88.0 & 69.0 \\
			\rowcolor{gray!25}Anti-Collapse\textcolor{red}{$^\star$} & IBN/64 & 63.1(+1.4) & 73.9 & 67.5 & 80.4(+1.6) & 88.2(+1.2) & 69.3 \\     
			\toprule
			\bottomrule
			HMP\textcolor{red}{$^\diamond$}\cite{wang2023hierarchical} $_\textrm{DSP2023}$        & IBN/128 & 68.3 & - & - & 85.9 & - & -    \\ 
			\hline
			\rowcolor{gray!25}Anti-Collapse                         & IBN/128 & 66.8 & 77.2 & 69.3 & 83.9 & 90.5 & 70.7     \\ 
			\rowcolor{gray!25}Anti-Collapse\textcolor{red}{$^\star$}           & IBN/128 & 67.7 & 77.9 & 70.4 & 84.3 & 90.3 & 70.5     \\
			\rowcolor{gray!25}Anti-Collapse\textcolor{red}{$^{\star\diamond}$} & IBN/128 & 69.4(+1.7) & 80.1 & 72.4 & 85.9 & 92.1 & 72.7     \\  
			\toprule
			\bottomrule
			ProxyAnchor ~\cite{Authors63}   $_\textrm{CVPR2020}$      & IBN/512 & 68.4 & 79.2 & - & 86.8 & 91.6 & -  \\ 
			ProxyGML   ~\cite{zhu2020fewer}   $_\textrm{NeurIPS2020}$                                & IBN/512 & 66.6 & 77.6 & 69.8 & 85.5 & 91.8 & 72.4   \\     
			PA + \textrm{MemVir}~\cite{ko2021learning}     $_\textrm{ICCV2021} $               & IBN/512 & 69.0 & 79.2 & - & 86.7 & 92.0 & -\\
			AHS \cite{yan2023adaptive}		   $_\textrm{TIP2023} $       & IBN/512 & 66.8 & 77.4 & - & 85.4 & 91.2 & - \\
			DFML \cite{wang2023deep}		  $_\textrm{CVPR2023} $        & IBN/512 & 69.3 & - & - & 88.4 & - & - \\
			Multi-Proxy \cite{chan2023multi}   $_\textrm{IS2023} $           & IBN/- & 69.6 & 79.9 & - & 90.3 & 93.7 & - \\
			\hline
			\rowcolor{gray!25}Anti-Collapse               & IBN/512 & 69.4 & 79.7 & 71.1 & 87.1  & 92.5 & 73.1 (+0.7) \\     
			\rowcolor{gray!25}Anti-Collapse\textcolor{red}{$^\star$} & IBN/512  & 70.8(+1.2) & 80.5(+0.6) & 71.4(+1.6) & 89.5(-0.8) & 93.2(-0.5) & 72.9 \\     
			\toprule
			\bottomrule
			mcSAP \cite{zhao2023multi}$_\textrm{NCA2023} $ & R50/64 & 60.5 & 72.4 & 68.7 & 78.9 & 87.1 & 70.9 \\
			\hline
			\rowcolor{gray!25}Anti-Collapse & R50/64 & 63.2 & 74.7 & 68.3 & 82.5 & 89.5 & 70.1 \\     
			\rowcolor{gray!25}Anti-Collapse\textcolor{red}{$^\star$} & R50/64  & 64.8(+4.3) & 76.1(+3.7) & 68.7 & 83.5(+4.6) & 89.9(+2.8) & 71.2(+0.3) \\     
			\toprule
			\bottomrule
			Div\&Conq\cite{sanakoyeu2019divide} $_\textrm{CVPR2019} $ & R50/128 & 65.9 & 76.6 & 69.6 & 84.6 & 90.7 & 70.3 \\
			MIC  ~\cite{roth2019mic}$_\textrm{ICCV2019} $ & R50/128 & 66.1 & 76.8 & 69.7 & 82.6 & 89.1 & 68.4 \\
			PADS ~\cite{roth2020pads} $_\textrm{CVPR2020} $  &R50/128 & 67.3 & 78.0 & 69.9 & 83.5 & 89.7 & 68.8 \\
			RankMI ~\cite{kemertas2020rankmi} $_\textrm{CVPR2020} $& R50/128 & 66.7 & 77.2 & 71.3 & 83.3 & 89.8 & 69.4 \\
			\hline
			\rowcolor{gray!25}Anti-Collapse               & R50/128 & 67.3 & 78.4 & 70.1 & 85.8 & 91.5 & 71.8 \\
			\rowcolor{gray!25}Anti-Collapse\textcolor{red}{$^\star$} & R50/128 & 68.3(+1.0) & 79.0(+1.0) & 72.5(+1.2) & 87.1(+2.5) & 92.5(+1.8) & 73.0(+2.7) \\   
			\toprule
			\bottomrule
			ProxyNCA++\cite{teh2020proxynca++}  $_\textrm{ECCV2020} $    & R50/1024 & 70.2 & 80.7 &  -   & 87.6 & 93.1 &  -  \\
			\hline 
			\rowcolor{gray!25}Anti-Collapse               & R50/1024 & 70.4 & 80.6 & 73.1 & 88.9 & 93.3 & 73.7 \\
			\rowcolor{gray!25}Anti-Collapse\textcolor{red}{$^\star$} & R50/1024 & 71.9(+1.7) & 81.7(+1.0) & 73.7 & 90.8(+1.9) & 94.7 (+1.4)& 75.5 \\     
			\toprule
			\bottomrule
			Circle Loss~\cite{sun2020circle}  $_\textrm{CVPR2020} $  & R50/512 & 66.7 & 77.2 & - & 83.4 & 89.7 & - \\
			DiVA~\cite{milbich2020diva}   $_\textrm{ECCV2020} $   & R50/512 & 69.2 & 79.3 & 71.4 & 87.6 & 92.9 & 72.2 \\
			DCML-MDW~\cite{zheng2021deep}$_\textrm{CVPR2021} $  & R50/512 & 68.4 & 77.9 & 71.8 & 85.2 & 91.8 & 73.9 \\
			PA+NIR~\cite{roth2022non}\textcolor{red}{$^\star$}  $_\textrm{CVPR2022} $    & R50/512 & 70.5& 80.6& 72.5  & 89.1 & 93.4 & 75.0 \\ 
			mcSAP \cite{zhao2023multi}$_\textrm{NCA2023} $& R50/512 & 63.5 & 75.6 & 71.0 & 84.6 & 91.5 & 74.3 \\
			IDML \cite{wang2023introspective}$_\textrm{TPAMI2023} $& R50/512 & 69.0 & 79.5 & 73.5 & 86.3 & 92.2 & 74.1\\
			HSE\cite{yang2023hse} $_\textrm{ICCV2023} $ & R50/512& 70.6 & 80.1 & - & 89.6 & 93.8 & - \\
			\hline 
			\rowcolor{gray!25}Anti-Collapse & R50/512 & 70.2  & 80.1  & 72.1  & 88.5  & 93.5 & 74.9 \\
			\rowcolor{gray!25}Anti-Collapse\textcolor{red}{$^\star$} & R50/512 & 71.7(+1.1) & 81.2(+0.6) & 73.2(-0.3) & 90.5(+0.9) & 94.6(+0.8) & 75.6(+0.6) \\      
			\toprule
	\end{tabular}}
	\caption{Performances Comparison of Anti-Collapse Loss with State-of-the-Art Methods on CUB200 and Cars196. \textcolor{red}{$\star$} indicates the use of double pooling operation; \textcolor{red}{$^\diamond$} is used to represent the use of large size input samples; - indicates that the data is not provided in the corresponding paper of the method; and the gray area denotes the experimental results of Anti-Collapse Loss.}
	\label{tab:sota}
\end{table*}

\begin{table}[h]
	\renewcommand{\arraystretch}{1.2}
	\setlength\tabcolsep{0.1pt}
	\footnotesize
	\centering
	\resizebox{1\linewidth}{!}{
		\begin{tabular}{l | c | c | c | c}
			\bottomrule
			\multicolumn{2}{c|}{Datasets} & \multicolumn{3}{c}{\textsc{SOP}}\\
			\toprule
			\bottomrule
			\textsc{Methods} & Arch/Dim. & R@1 & R@10 & NMI\\
			\toprule
			\bottomrule
			PNCA~\cite{Authors55} $_\textrm{ICCV2017}$        & IBN/64 & 73.7 & 89.0 & -   \\
			MS~\cite{Authors7}$_\textrm{CVPR2019}$             & IBN/64 & 74.1 & 87.8 & -   \\
			SoftTriple~\cite{qian2019softtriple} $_\textrm{ICCV2019}$  & IBN/64 & 76.3 & 89.1 & -   \\ 
			PA ~\cite{Authors63}$_\textrm{CVPR2020}$ & IBN/64 & 76.5 & 89.0 & -   \\ 
			\hline
			\rowcolor{gray!25}Anti-Collapse               & IBN/64 & 76.3 & 89.1 & 89.8 \\
			\rowcolor{gray!25}Anti-Collapse\textcolor{red}{$^\star$} & IBN/64 & 77.0(+0.5) & 89.4(+0.4) & 90.1 \\     
			\toprule
			\bottomrule
			HMP\textcolor{red}{$^\diamond$}\cite{wang2023hierarchical} $_\textrm{DSP2023}$        & IBN/128 & 75.8 & - & -   \\ 
			\hline
			\rowcolor{gray!25}Anti-Collapse                         & IBN/128 & 74.4 & 85.2 & 89.5   \\ 
			\rowcolor{gray!25}Anti-Collapse\textcolor{red}{$^\star$}           & IBN/128 & 74.7 & 85.7 & 88.7   \\
			\rowcolor{gray!25}Anti-Collapse\textcolor{red}{$^{\star\diamond}$} & IBN/128 & 75.9(+0.1) & 86.1 & 88.9   \\  
			\toprule
			\bottomrule
			PA ~\cite{Authors63}   $_\textrm{CVPR2020}$      & IBN/512 & 79.1 & 90.8 & -   \\ 
			PGML   ~\cite{zhu2020fewer}   $_\textrm{NeurIPS2020}$                                & IBN/512 & 78.0 & 90.6 & 90.2   \\     
			PA$+$\textrm{MemVir}~\cite{ko2021learning}     $_\textrm{ICCV2021} $               & IBN/512 & 79.7 & 91.0 & - \\
			AHS \cite{yan2023adaptive}		   $_\textrm{TIP2023} $       & IBN/512 & 78.5 & 90.7 & - \\
			Multi-Proxy \cite{chan2023multi}   $_\textrm{IS2023} $           & IBN/- & 80.1 & 91.3 & - \\
			\hline
			\rowcolor{gray!25}Anti-Collapse               & IBN/512 & 79.4 & 90.5 & 90.3  (+0.1)\\     
			\rowcolor{gray!25}Anti-Collapse\textcolor{red}{$^\star$} & IBN/512  & 80.1 & 91.1(-0.2) & 90.1 \\     
			\toprule
			\bottomrule
			mcSAP \cite{zhao2023multi}$_\textrm{NCA2023} $ & R50/64 & 77.3 & 89.8 & 91.7 \\
			\hline
			\rowcolor{gray!25}Anti-Collapse & R50/64 & 77.8 & 89.6 & 90.2(-1.5) \\     
			\rowcolor{gray!25}Anti-Collapse\textcolor{red}{$^\star$} & R50/64 & 78.1(+0.8) & 90.2(+0.4) & 89.2 \\     
			\toprule
			\bottomrule
			Div\&Conq\cite{sanakoyeu2019divide} $_\textrm{CVPR2019} $ & R50/128 & 75.9 & 88.4 & 90.2\\
			MIC  ~\cite{roth2019mic}$_\textrm{ICCV2019} $ & R50/128 & 77.2 & 89.4 & 90.0\\
			PADS ~\cite{roth2020pads} $_\textrm{CVPR2020} $  &R50/128 & 76.5 & 89.0 & 89.9\\
			RankMI ~\cite{kemertas2020rankmi} $_\textrm{CVPR2020} $& R50/128 & 74.3 & 87.9 & 90.5 \\
			\hline
			\rowcolor{gray!25}Anti-Collapse               & R50/128& 79.7 & 91.1(+1.7) & 90.7(+0.2)\\
			\rowcolor{gray!25}Anti-Collapse\textcolor{red}{$^\star$} & R50/128 & 79.8 (+2.6)& 91.0 & 90.3 \\   
			\toprule
			\bottomrule
			PNCA$++$\cite{teh2020proxynca++}  $_\textrm{ECCV2020} $    & R50/1024 & 80.7 & 92.0 &  -\\
			\hline 
			\rowcolor{gray!25}Anti-Collapse               & R50/1024 & 80.3 & 91.2 & 89.9\\
			\rowcolor{gray!25}Anti-Collapse\textcolor{red}{$^\star$} & R50/1024 & 81.2(+0.5) & 92.0 & 90.4 \\     
			\toprule
			\bottomrule
			Circle Loss~\cite{sun2020circle}  $_\textrm{CVPR2020} $  & R50/512 & 78.3 & 90.5 & - \\
			DiVA~\cite{milbich2020diva}   $_\textrm{ECCV2020} $   & R50/512 & 79.6 & 91.2 & 90.6 \\
			DCML-MDW~\cite{zheng2021deep}$_\textrm{CVPR2021} $  & R50/512 & 79.8 & 90.8 & 90.8 \\
			PA+NIR~\cite{roth2022non}\textcolor{red}{$^\star$}  $_\textrm{CVPR2022} $    & R50/512 & 80.4 & 91.4 & 90.6 \\ 
			mcSAP \cite{zhao2023multi}$_\textrm{NCA2023} $& R50/512 & 79.9 & 91.5 & 92.2 \\
			IDML \cite{wang2023introspective}$_\textrm{TPAMI2023} $& R50/512 & 79.7 & 91.4 & 91.2 \\
			HSE\cite{yang2023hse} $_\textrm{ICCV2023} $ & R50/512 & 80.0 & 91.4 & - \\
			\hline 
			\rowcolor{gray!25}Anti-Collapse & R50/512 & 81.2(+0.8) & 92.0(+0.5) & 91.5  \\
			\rowcolor{gray!25}Anti-Collapse\textcolor{red}{$^\star$} & R50/512 & 81.0 & 91.6 & 91.6(-0.6) \\      
			\toprule
	\end{tabular}}
	\caption{Performances Comparison of Anti-Collapse Loss with State-of-the-Art Methods on SOP. \textcolor{red}{$\star$} indicates the use of double pooling operation; \textcolor{red}{$^\diamond$} is used to represent the use of large size input samples; - indicates that the data is not provided in the corresponding paper of the method; and the gray area denotes the experimental results of Anti-Collapse Loss.}
	\label{tab:sota1}
\end{table}

\begin{figure}[t]
	\centering
	\includegraphics[width=0.98\linewidth]{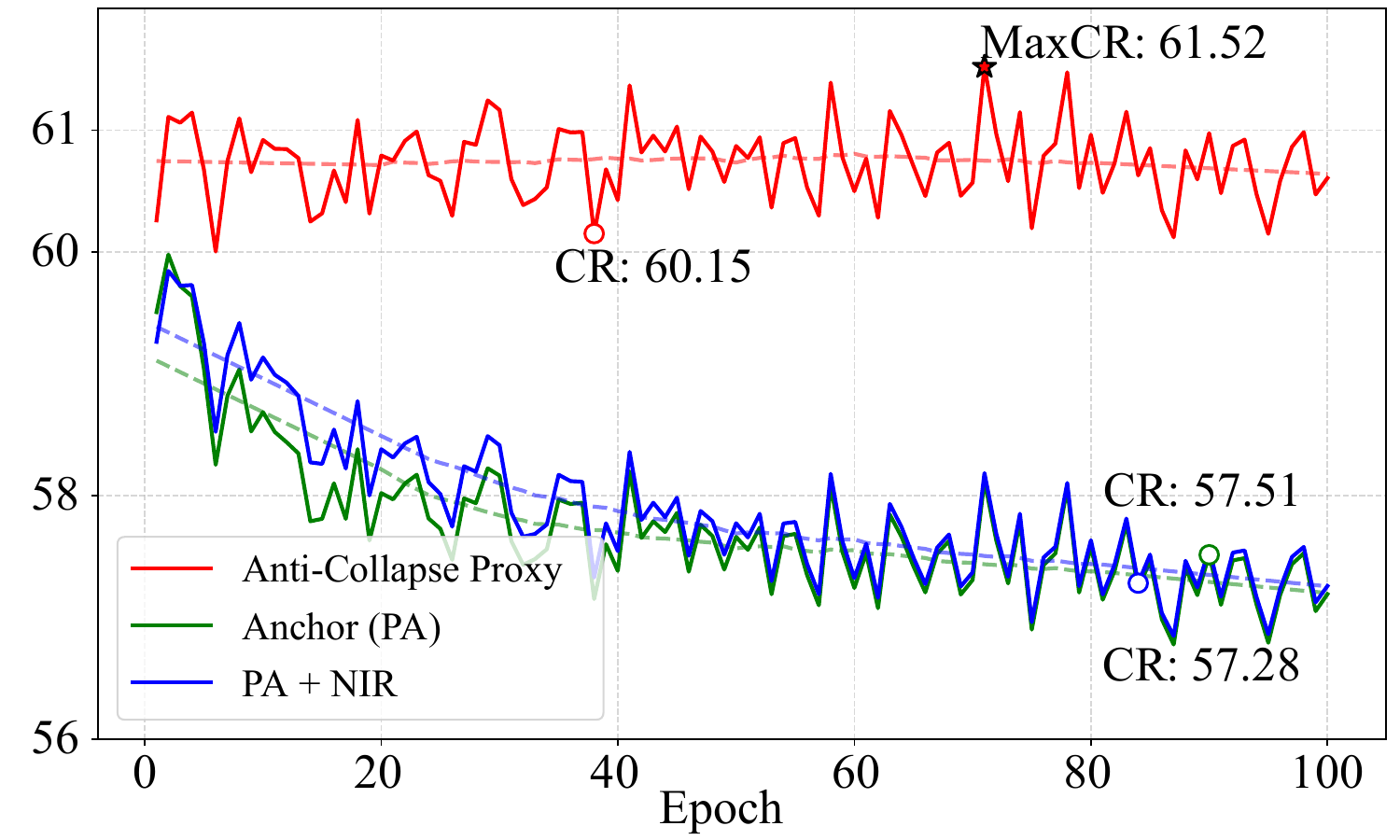}
	\caption{The coding rate variation of three proxy-based losses during training.}
	\label{Fig:pxy_cr}
\end{figure}

\begin{table}[b]
	\renewcommand{\arraystretch}{1.2}
	\centering
	\resizebox{1\linewidth}{!}{
		\begin{tabular}{l | c | c | c | c }
			\bottomrule
			\textbf{Methods} & $R_{global}$ & $R_{intra}$ & $R_{proxy}$ & $\rho_{density}$   \\
			\toprule
			\bottomrule
			PA~\cite{Authors63} $_\textrm{CVPR2020} $                   & 150.0 & 60.0 & 57.5   & 0.71           \\
			PA+NIR  ~\cite{roth2022non} $_\textrm{CVPR2022} $                 & 147.4 & 59.2 & 57.2  & 0.70         \\
			\rowcolor{gray!25}Anti-Collapse $(\textbf{ours})$           & \textbf{119.5} & \textbf{52.5} & \textbf{60.1}  & \textbf{0.69}  \\
			\toprule
	\end{tabular}}
	\caption{Comparison of Embedding Space Structural Metrics. We test four metrics that can measure the structure of embedding space on the CUB200 dataset. The backbone network used is BN-Inception, and the embedding dimension is set to 512.}
	\label{tab:emb_exp}
\end{table}

\subsection{Experimental Results} 

We select a variety of mapping dimensions for testing, allowing us to conduct fair comparisons with other methods. The experimental results are present in Table \ref{tab:sota} and Table \ref{tab:sota1}. It's worth mentioning that the use of dual pooling operation in the backbone network is quite common in image retrieval tasks based on deep metric learning, but some works do not explicitly state this. Hence, we could not add the \textcolor{red}{$\star$} mark to some of the methods. From Table \ref{tab:sota} and Table \ref{tab:sota1}, we can observe that the Anti-Collapse Loss achieves the best recall rate (Recall@1(\%)) and normalized mutual information (NMI) on each dataset. We first analyze the part of the experiment with ResNet50 acting as the backbone network. When the feature dimension is 512, the recall results (\%) of Anti-Collapse Loss are 71.7 VS 70.6 (+1.1) on CUB200, 90.5 VS 89.6 (+0.9) on Cars196, and 81.2 VS 80.4 (+0.8) on SOP. Under the setting where the feature dimension is 128, our method results are 68.3 VS 67.3 (+1.0) on CUB200, 87.1 VS 83.5 (+2.5) on Cars196, and 79.8 VS 77.2 (+2.6) on SOP. When the network is BN-Inception, and the feature dimension is 512, our experimental results on CUB200, Cars196, and SOP are 70.8 VS 69.6 (+1.2), 89.5 VS 90.3 (-0.8), and 80.1 VS 80.1, respectively. The above descriptions are the experimental results under the experimental settings that are more frequently used in existing deep metric learning methods. Besides, we can also notice the competitive image retrieval performance of Anti-Collapse Loss under other settings in Table \ref{tab:sota} and Table \ref{tab:sota1}.

\begin{figure*}[t]
	\centering
	\subcaptionbox{\label{fig:start_hist}{Initial.}}{
		\includegraphics[width=0.23\linewidth, trim={1.6cm 0.4cm 0.2cm 1.0cm}, clip]{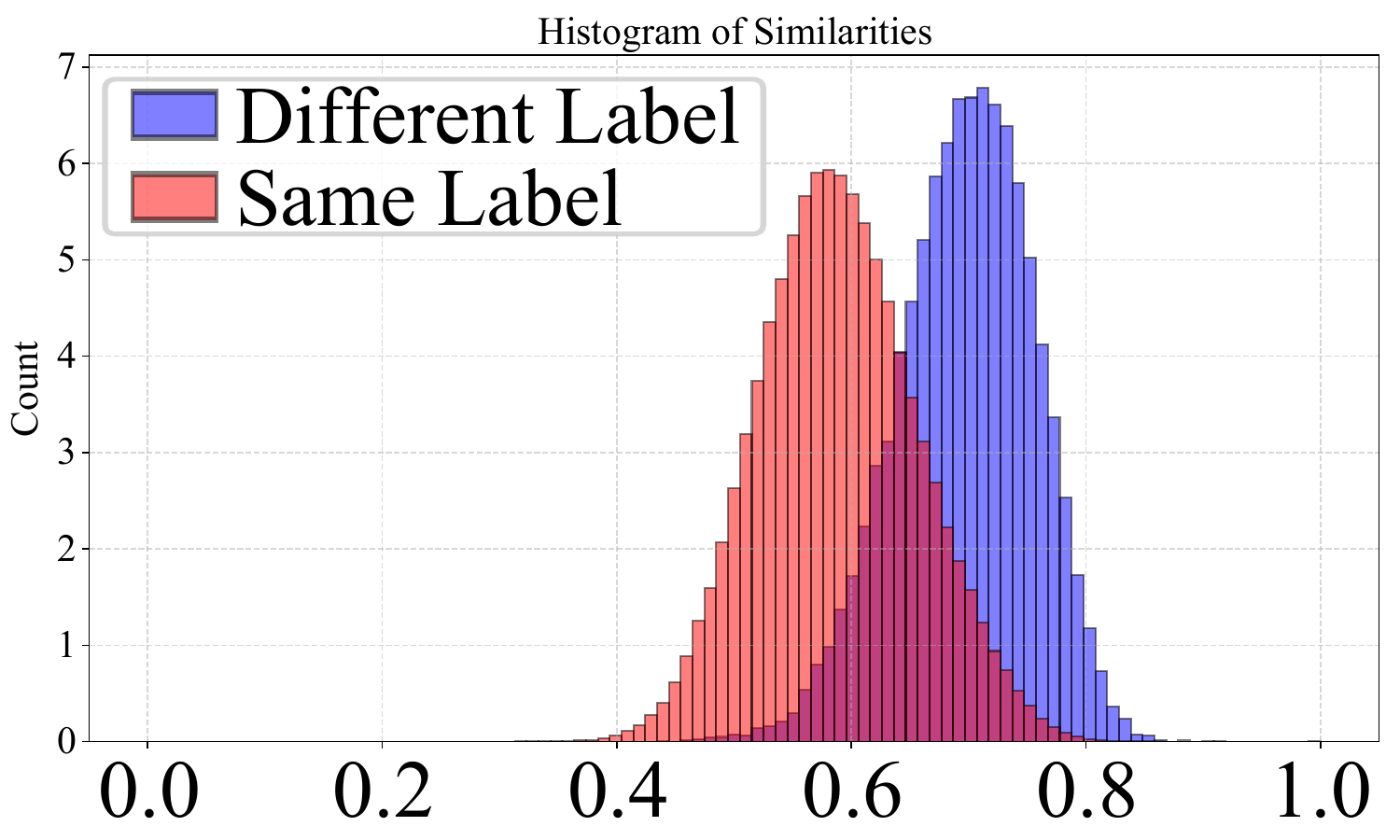}
	}
	\subcaptionbox{\label{fig:pa-best_hist}{PA}}{
		\includegraphics[width=0.23\linewidth, trim={1.6cm 0.4cm 0.2cm 1.0cm}, clip]{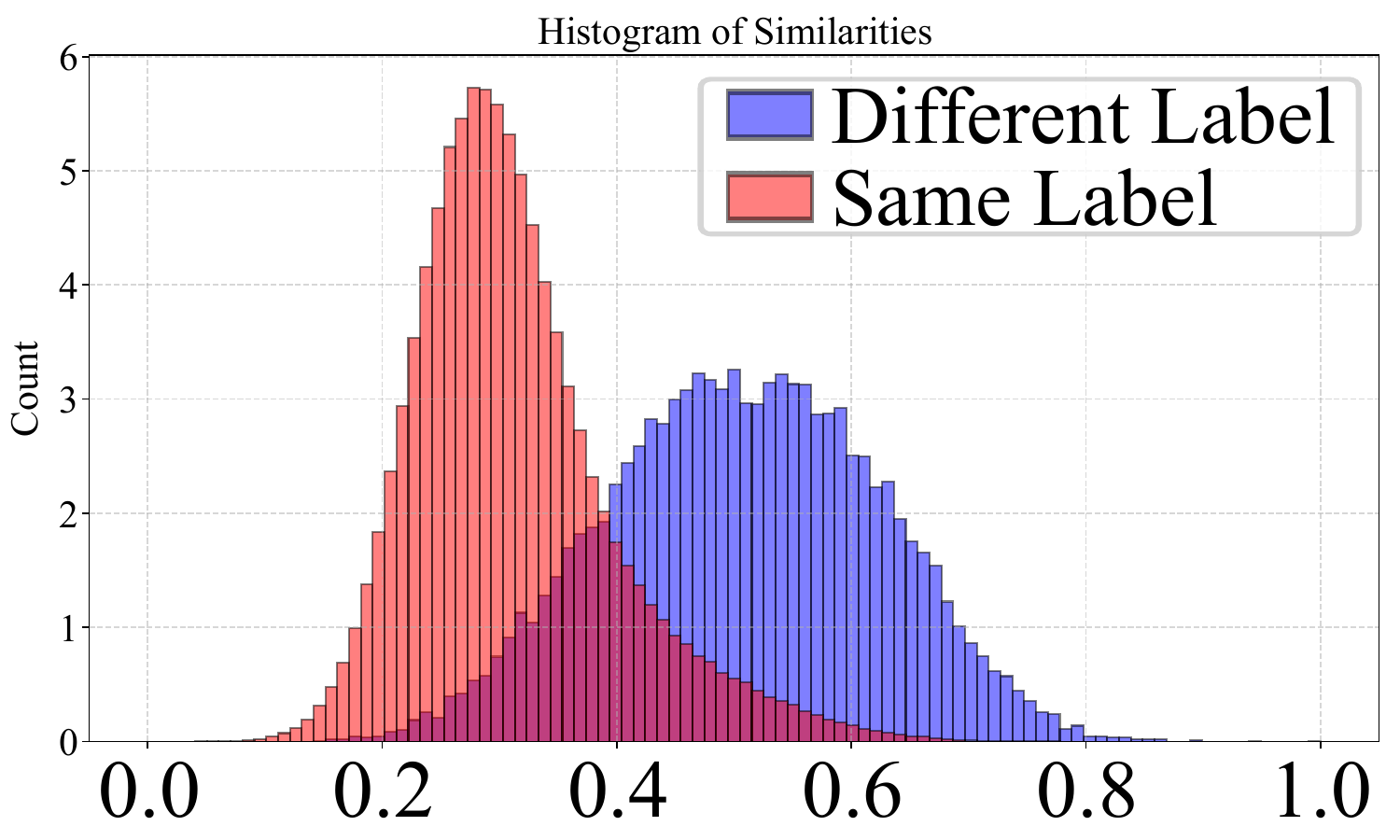}
	}
	\subcaptionbox{\label{fig:nir-best_hist}{NIR}}{
		\includegraphics[width=0.23\linewidth, trim={1.6cm 0.4cm 0.2cm 1.0cm}, clip]{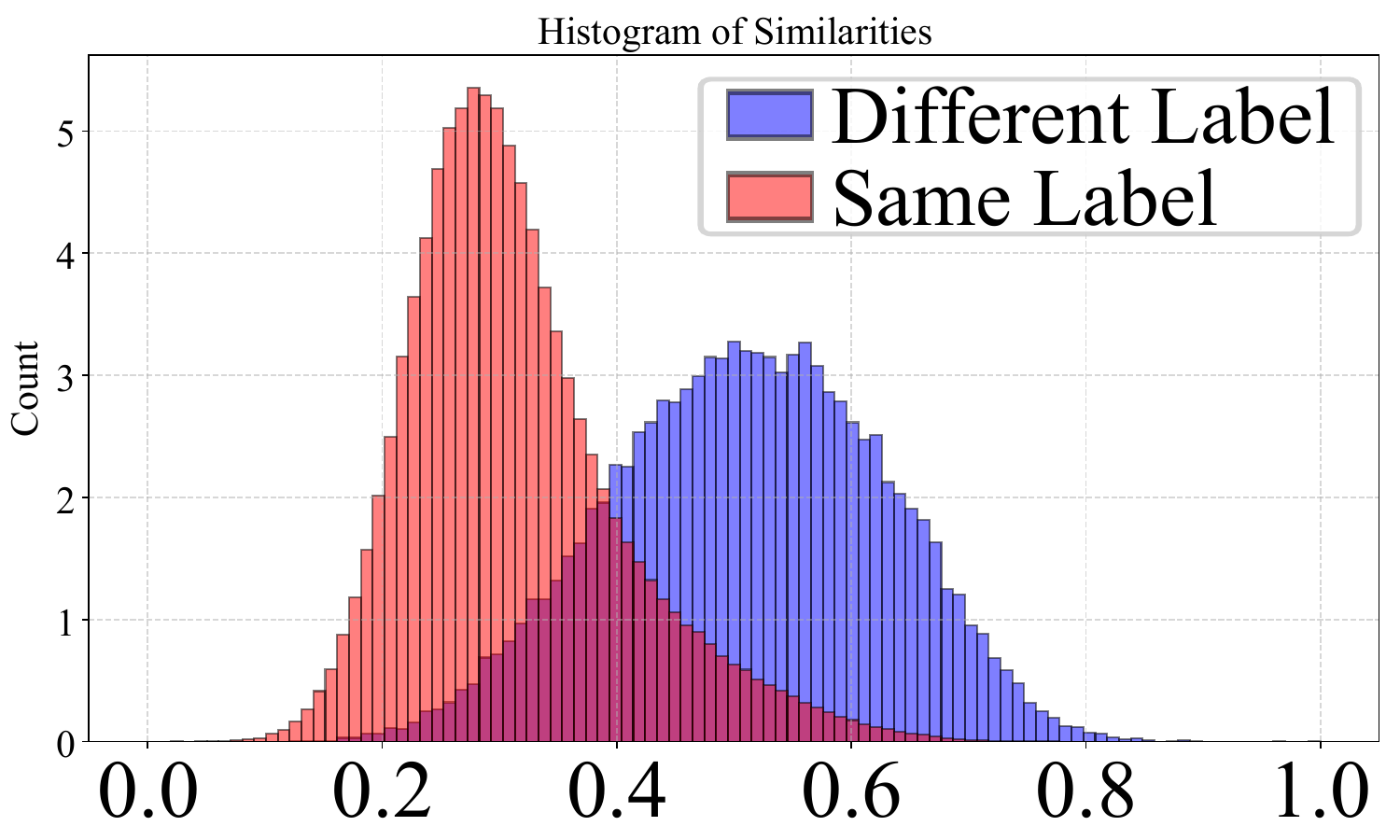}
	}
	\subcaptionbox{\label{fig:anti-collapse-best_hist}{\textbf{Ours}}}{
		\includegraphics[width=0.23\linewidth, trim={1.6cm 0.4cm 0.2cm 1.0cm}, clip]{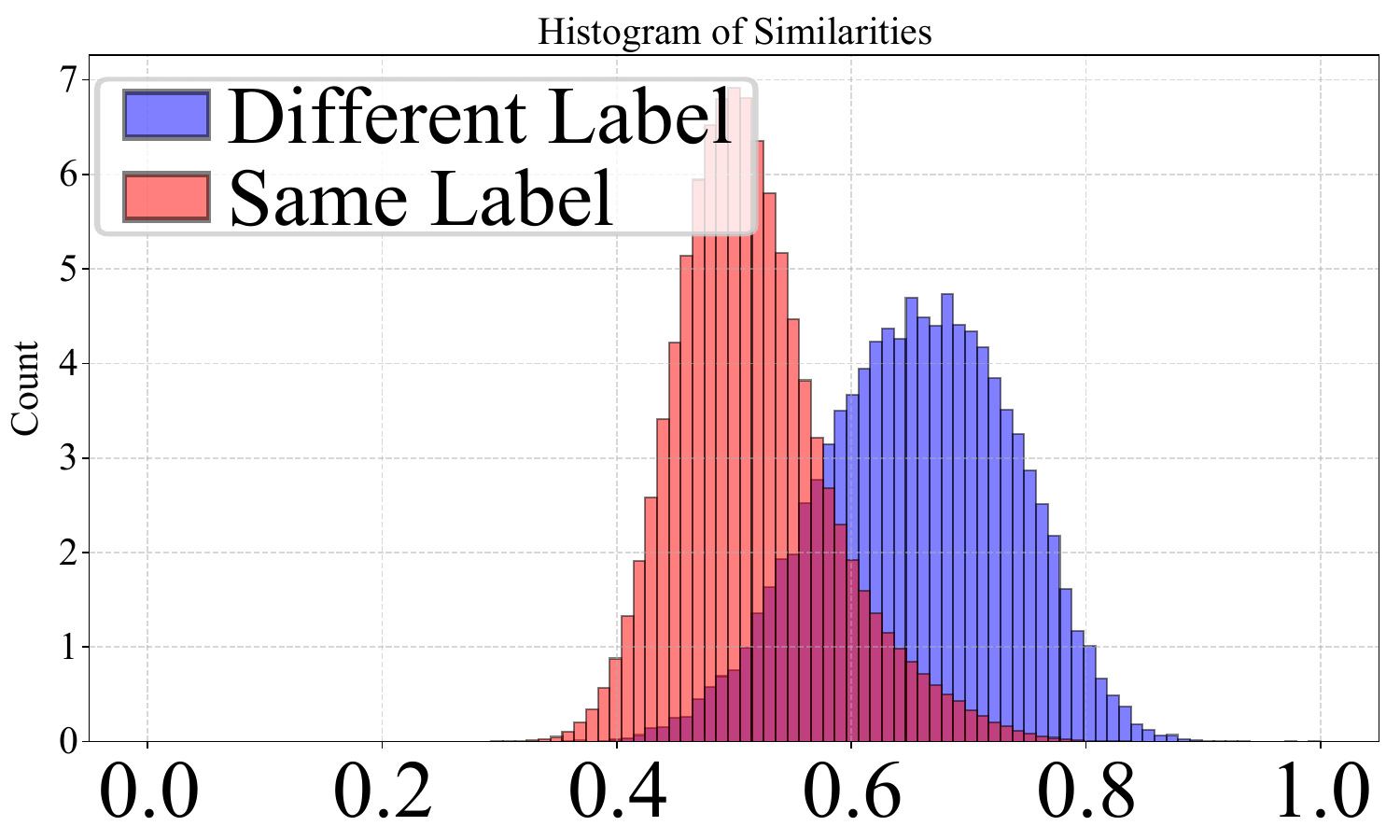}
	}
	\caption{The histogram of sample similarity distribution for proxy-based loss in achieving optimal image retrieval performance (Max Recall@1).}
	\label{fig:all-paircol}
\end{figure*}

\begin{figure}[t]
	\centering
	
	\subcaptionbox{\label{fig:start_hm}{Initial.}}{
		\includegraphics[width=0.2\linewidth, trim={2.5cm 1.2cm 4cm 1.2cm}, clip]{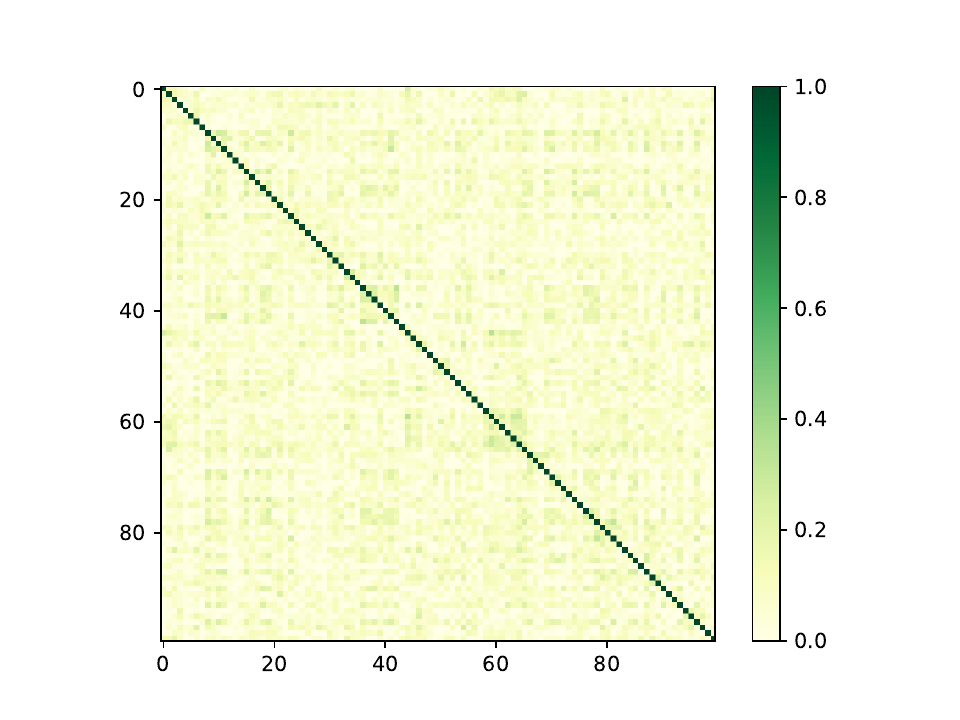}
	}
	\subcaptionbox{\label{fig:pa-best_hm}{PA}}{
		\includegraphics[width=0.2\linewidth, trim={2.5cm 1.2cm 4cm 1.3cm}, clip]{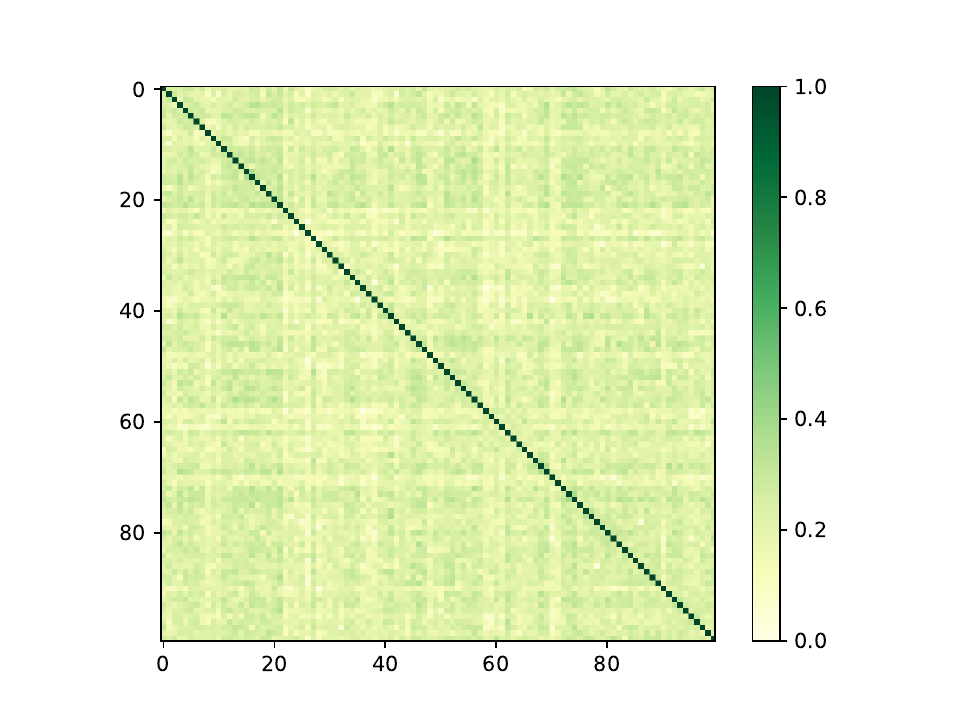}
	}
	\subcaptionbox{\label{fig:nir-best_hm}{NIR}}{
		\includegraphics[width=0.2\linewidth, trim={2.5cm 1.2cm 4cm 1.3cm}, clip]{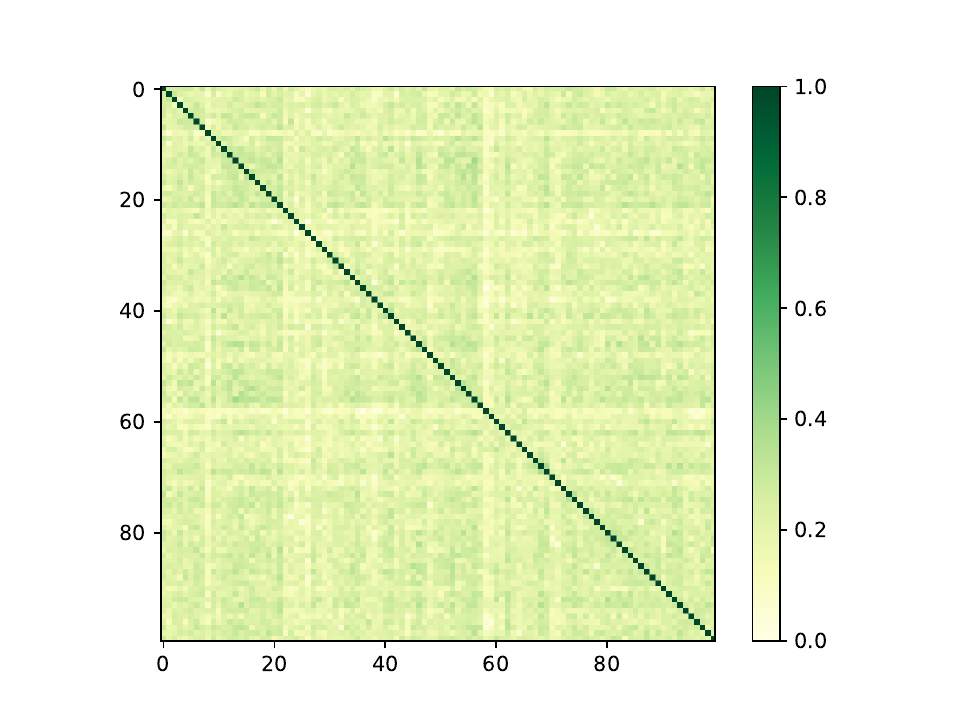}
	}
	\subcaptionbox{\label{fig:anti-collapse-best_hm}{\textbf{Ours}}}{
		\includegraphics[width=0.2\linewidth, trim={2.5cm 1.2cm 4cm 1.3cm}, clip]{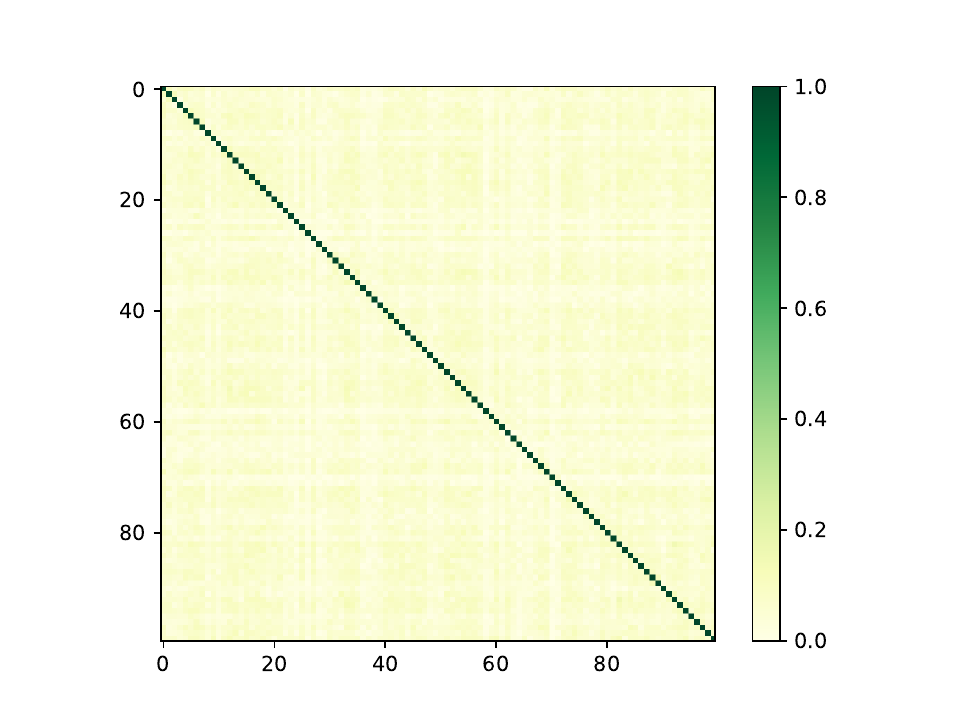}
	}
	\caption{Proxy similarity matrix of different proxy-based losses in achieving optimal image retrieval performance.}
	\label{fig:all-proxysim}
\end{figure}

\subsection{Anti-Collapse and Convergence Performances}            

In addition to demonstrating the performance of our method in image retrieval, we also verify its ability to optimize the embedding space. To this end, we utilize three code rate metrics to evaluate the performance of the Anti-Collapse loss, including the global Gaussian coding rate $R_{global}$~\cite{yu2020learning}, the intra-class coding rate $R_{intra}$, and the proxy coding rate $R_{proxy}$. Additionally, we also use the embedding space density $\rho_{density} = \frac{\rho_{intra}(d_{pos})}{\rho_{inter}(d_{neg})}$ \cite{roth2020revisiting} as one of our evaluation indicators. Here, $\rho_{intra}$ represents the average intra-class distance, and $\rho_{inter}$ represents the average inter-class distance. 
A larger value of $\rho_{density}$ indicates a more sparse distribution of intra-class samples and a more compact distribution of inter-class samples, indicating better generalization performance and embedding space structure.
It is important to note that $\rho_{density}$ is a relative parameter with certain limitations. When the model is only loaded but not yet trained, dispersed samples can also result in a higher embedding space density. Therefore, we use the embedding space density $\rho_{density}$ when achieving the best recall rate Recall@$1$ (\%). From Table \ref{tab:emb_exp}, we can observe that in both ProxyAnchor~\cite{Authors63} and NIR~\cite{roth2020revisiting} methods, the class proxies experience collapse in the embedding space, i.e., the coding rate of proxies decreases when achieving the maximum recall rate. In contrast, our Anti-Collapse method maintains a higher coding rate. 

To more comprehensively present the changes in the embedding space, we record the variations in the coding rate of the proxies throughout the entire training process. The experimental results are given in Fig.~\ref{Fig:pxy_cr}. By observing Fig.~\ref{Fig:pxy_cr}, we can notice that Anti-Collapse effectively prevents the collapse of proxies in the embedding space. It ensures that the coding rate of the proxy remains relatively stable throughout the entire training process, with a mean of 60.74 and a variance of 0.1. In contrast, the coding rates of the other two methods continuously decrease, indicating the collapse of the proxy set in the embedding space. Additionally, by observing the time it takes for each method to achieve optimal recall rate, we can see that Anti-Collapse achieves the best retrieval results as early as the 38th epoch, while ProxyAnchor Loss and NIR Loss achieve their best recall rates at the 83rd and 90th epochs, respectively. This experimental result demonstrates the excellent ability of Anti-Collapse to accelerate convergence. 

The experimental results of the distribution for similarities between samples are present in Fig.~\ref{fig:all-paircol}. From Fig.~\ref{fig:all-paircol}, we can notice that when achieving optimal retrieval performance, the Anti-Collapse loss exhibits a more compact distribution of intra-class and inter-class similarities than the other two methods.

Fig.~\ref{fig:all-proxysim} presents the similarity matrix of proxies. Specifically, Fig.~\ref{fig:all-proxysim} (a) displays the similarity between proxies during initialization, while the other three sub-figures show the similarity between proxies for different methods when achieving maximum Recall@1. The color is darker in the similarity graphs of Fig.~\ref{fig:all-proxysim} (b) and Fig.~\ref{fig:all-proxysim} (c), indicating that the algorithms of NIR and PA have insufficient inter-class discrimination capability. Our Anti-Collapse Loss encourages proxies to be as orthogonal as possible in the embedding space (demonstrated in Fig.~\ref{fig:all-proxysim} (d)), enabling our approach to achieve better classification capabilities.

\subsection{Transfer Performance}

Generalization performance is an important aspect of the ability of deep metric learning, which involves representation learning. One significant objective of our proposed Anti-Collapse Loss is to prevent the model from overfitting to the training set by maintaining the structure of the embedding space. To this end, we conduct generalization performance experiments on the Anti-Collapse Loss and compare it with some existing methods with good image retrieval performance.

\begin{table}[b]
	\footnotesize
	\centering
	\renewcommand{\arraystretch}{1.2}
	\resizebox{1\linewidth}{!}{
		\begin{tabular}{l | c | c | c | c | c | c}
			\bottomrule
			\multicolumn{1}{c|}{Datasets} & \multicolumn{2}{c|}{NABirds-A} & \multicolumn{2}{c|}{NABirds-B} & \multicolumn{2}{c}{NABirds-C}\\
			\toprule
			\bottomrule
			\textsc{Methods}  & R@1 & R@2 & R@1 & R@2  & R@1 & R@2 \\
			\toprule
			\bottomrule
			ProxyAnchor (PA)  & 73.1 & 81.7 & 68.1 & 78.2  & 59.2 & 69.3  \\ 
			PA+NIR   & 73.5 & 82.3 & 70.1 & 79.2 & 60.8 & 71.3   \\
			\hline
			\rowcolor{gray!25}Anti-Collapse & 74.7 & 83.1 & 71.7 & 80.6  & 62.5 & 72.6 \\     
			\toprule
	\end{tabular}}
	\caption{Results (Recall@$k$) of the generalization performance comparison between the Anti-Collapse Loss and existing methods.}
	\label{tab:trans}
\end{table}

\begin{figure}[t]
	\centering
	\includegraphics[width=\linewidth]{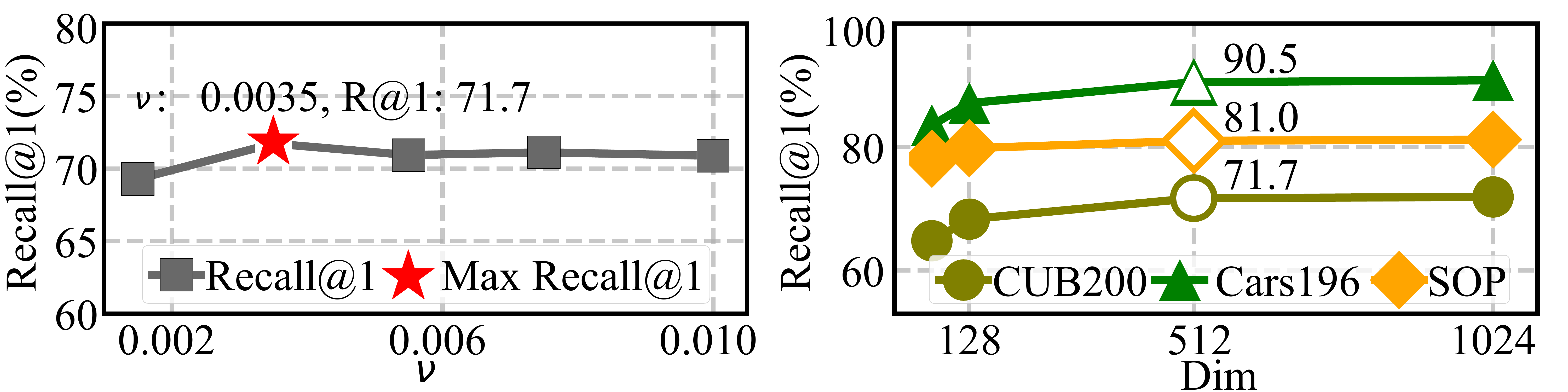}
	\caption{The impact of parameters $\nu$ and embedding dimension on Recall@1.}
	\vspace{-0.2cm}
	\label{Fig:nu_dim}
	\vspace{-0.3cm}
\end{figure}

To be specific, we still choose the first 100 classes of the CUB200 dataset for this experiment as the training set. As for the test set, we select another commonly used fine-grained bird dataset called NABirds~\cite{van2015building}. This dataset comprises a total of 48,562 images belonging to 555 bird species. We divide them into three test sets based on category indexes, each containing 200, 200, and 155 classes of bird images, respectively.
In the experiment, we use ResNet50 as the backbone network for training, with an embedding dimension of 512 and a batch size of 90. As shown in Table \ref{tab:trans}, we observe that the Anti-Collapse Loss achieves the best retrieval performance on all three test sets. For the retrieval metric Recall@1(\%), our proposed Anti-Collapse loss outperforms PA+NIR~\cite{roth2022non} by 1.2, 1.6, and 1.7, respectively. These results demonstrate that the Anti-Collapse Loss enables the model to achieve better generalization performance by continuously maintaining the structure of the embedding space during training.

\subsection{Ablation Studies}

\textbf{Parameter Sensitivities}: We investigate the impact of parameters $\nu$ and embedding dimension on the retrieval performance of Anti-Collapse Losses. Fig.~\ref{Fig:nu_dim} in the left subtable demonstrates the influence of different values of $\nu$ on the image retrieval performance in the CUB200 dataset. Our algorithm achieves the highest recall rate of 71.7 when $\nu$ is set to 0.0035. Furthermore, by examining the right subtable in Fig.~\ref{Fig:nu_dim}, we observe that as the dimension increases, the effect of retrieval performance gradually diminishes across all three datasets, with the algorithm's performance reaching a plateau at around 1024 dimensions.

\begin{table}[t]	
	\setlength\tabcolsep{1.5pt}
	\renewcommand{\arraystretch}{1.2}
	\footnotesize
	\centering
	\resizebox{\columnwidth}{!}{
		\begin{tabular}{l | c| c | c | c | c }			
			\bottomrule
			\rowcolor{gray!25}\multicolumn{6}{c}{CUB200}\\
			\toprule
			\bottomrule
			\textsc{Measurement index} & Unsup.&R@1 & F1 & NMI & mAP1000 \\
			\toprule
			\bottomrule
			ImageNet~\cite{russakovsky2015imagenet} &-& 53.3  & 27.1  & 60.4  & 20.1   \\
			$\mathcal{L}_{ProxyAnchor(PA)}$ & $\times$ &69.7 & - & - & - \\
			\rowcolor{gray!25}$	\mathcal{L}^{pair}(Eq.(\ref{eq:ac_pair}))$ & $\surd$ & 59.8  & 30.0  & 62.1  &23.8 \\
			\rowcolor{gray!25}$	\mathcal{L}^{pair}(Eq.(\ref{eq:ac_pair}))+\mathcal{L}_{proxy} $   &$\times$ & 70.9  & 42.3  & 72.1  & 35.9   \\
			\rowcolor{gray!25}$	\mathcal{L}_{AntiCo}^{proxy (All-Class)} $  & $\times$ &71.1 & 45.2 & 73.3 & 36.7 \\  
			\rowcolor{gray!25}$	\mathcal{L}_{AntiCo}^{proxy (Mini-Batch)} $  & $\times$ &71.7 & 46.1 & 73.2 & 36.5 \\      
			\toprule
			\bottomrule
			\rowcolor{gray!25}\multicolumn{6}{c}{Cars196}\\
			\toprule
			\bottomrule
			\textsc{Measurement index} & Unsup.&R@1 & F1 & NMI & mAP1000 \\
			\toprule
			\bottomrule
			ImageNet ~\cite{russakovsky2015imagenet}&-& 43.4 & 11.2 & 40.6 & 7.8 \\
			$\mathcal{L}_{ProxyAnchor(PA)}$ &$\times$& 87.7 & - & - & - \\
			\rowcolor{gray!25}$	\mathcal{L}^{pair}(Eq.(\ref{eq:ac_pair}))$ & $\surd$ & 46.2  & 11.2  & 41.0  &8.4 \\
			\rowcolor{gray!25}$	\mathcal{L}^{pair}(Eq.(\ref{eq:ac_pair}))+\mathcal{L}_{proxy} $  &$\times$ & 89.9  & 47.2  & 75.2  & 38.3  \\
			\rowcolor{gray!25}$	\mathcal{L}_{AntiCo}^{proxy (All-Class)}$     & $\times$ &90.1 & 46.9 & 75.5 & 37.9 \\
			\rowcolor{gray!25}$	\mathcal{L}_{AntiCo}^{proxy (Mini-Batch)} $     & $\times$ &90.5 & 47.9 & 75.6 & 37.7 \\      
			\toprule
	\end{tabular}}
	\caption{Results of performance comparison among different Anti-Collapse Loss Terms on CUB200 and Cars196 datasets. Evaluation metrics include Recall@1(\%), F1(\%), NMI(\%), and mAP1000(\%).}
	\label{tab:as}
\end{table}

\textbf{Different Anti-Collapse Loss Terms:} For performance comparison among different Anti-Collapse Loss Terms, we use ResNet50 as the backbone and set the embedding dimension to 512. We record the results of Anti-Collapse Losses in the CUB200 and Cars196 datasets in Table \ref{tab:as}. 
From Table \ref{tab:as}, we can notice that when using only the pairwise Anti-Collapse Loss, the model's improvement in terms of R@1(\%) on two datasets is as follows: 59.8 VS 53.3 (+6.5) and 46.2 VS 43.4 (+2.8). Regarding the proxy-based Anti-Collapse loss, we present the effectiveness of proxy-based loss when it is not used. 

Additionally, we present commonly used metrics for classification clustering analysis. By observing Table \ref{tab:as}, we can find that Eq.(\ref{eq:ac_pair}) possesses self-supervised training capability. Existing metric learning losses exhibit significant performance improvement when combined with it. By comparing the two proxy-based Anti-Collapse Loss approaches, we observe that the version utilizing only the labels corresponding to the samples in the mini-batch achieves the best performance among the three Anti-Collapse Loss methods. It achieves R@1(\%) of 71.7 and 90.5 on CUB200 and Cars196 datasets, respectively. This also indicates that optimizing proxies in a more targeted manner can yield better gains in image retrieval performance.
\begin{table}[t]	
	\setlength\tabcolsep{1.5pt}
	\renewcommand{\arraystretch}{1.2}
	\footnotesize
	\centering
	\resizebox{\columnwidth}{!}{
		\begin{tabular}{l | c| c | c | c }			
			\bottomrule
			\rowcolor{gray!25}\multicolumn{5}{c}{CUB200}\\
			\toprule
			\bottomrule
			\textsc{Measurement index} &R@1 & F1 & NMI & mAP1000 \\
			\toprule
			\bottomrule
            $\mathcal{L}_{ProxyAnchor(PA)}$ & 69.7 & - & - & - \\
            $\mathcal{L}_{PA} + CLIP_{ViT-B/32}$  & 70.0 & 44.2 & 71.2 & 34.7 \\
            \rowcolor{gray!25}$	\mathcal{L}_{AntiCo} $  &71.7 & 46.3 & 73.2 & 36.5 \\
			\rowcolor{gray!25}$	\mathcal{L}_{AntiCo} + CLIP_{ViT-B/32}$  &72.0 & 46.3 & 73.6 & 36.9 \\      
			\toprule
			\bottomrule
			\rowcolor{gray!25}\multicolumn{5}{c}{Cars196}\\
			\toprule
			\bottomrule
			\textsc{Measurement index} &R@1 & F1 & NMI & mAP1000 \\
			\toprule
			\bottomrule
            $\mathcal{L}_{ProxyAnchor(PA)}$ & 87.7 & - & - & - \\
            $\mathcal{L}_{PA} + CLIP_{ViT-B/32}$  &88.2 & 45.2 & 73.6 & 35.1 \\
			\rowcolor{gray!25}$	\mathcal{L}_{AntiCo}  $  &90.5 & 47.9 & 75.6 & 37.7 \\    
            \rowcolor{gray!25}$	\mathcal{L}_{AntiCo} + CLIP_{ViT-B/32}$  & 90.9 & 48.2 & 76.0 & 37.9 \\
			\toprule
	\end{tabular}}
	\caption{Performance of deep metric learning methods integrated with large pre-trained vision-language model. Evaluation metrics include Recall@1(\%), F1(\%), NMI(\%), and mAP1000(\%).}
	\label{tab:lang_mod}
\end{table}
\subsection{Integrated Vision-Language Models Experiment}
Recently, methods based on large pre-trained vision-language models have achieved remarkable results in various visual classification tasks. This part attempts to integrate large pre-trained vision-language models with our proposed Anti-Collapse Loss. Our experiments utilize pre-trained language models to convert class labels into language embeddings. By aligning visual and language similarity matrices, we guide the learning of visual embedding spaces to enhance the semantic consistency and generalization ability of deep metric learning.
Specifically, we combine the class labels in the training set with a simple prompt template, ``a photo of \{label\}" to describe each image class. These sentences are then mapped into the language embedding space using a large pre-trained language model. Consequently, each class in the training set is assigned a corresponding textual feature. We establish higher-level semantic relationships between samples in mini-batches by calculating a similarity matrix through their label text features. The size of this similarity matrix matches that of the visual similarity matrix, which is $n_{bs}^2$. We then optimize the KL-Divergence between the language and visual similarity matrices to leverage language similarities for guiding visual metric learning. We employ CLIP's~\cite{radford2021learning} text encoder (ViT-B/32) as the backbone network for generating textual features, while ResNet50 serves as the backbone network for extracting image features. The features are normalized (L2  regularization) to the unit hypersphere after being output. We test the performance metrics of CLIP$+$AntiCo Loss and CLIP$+$ProxyAnchor combinations on CUB200 and Cars196. The results, presented in Table \ref{tab:lang_mod}, show that after integrating CLIP,  our proposed method's Recall@1 improves by 0.3\% on CUB200 and by 0.4\% on Cars196. These experimental results demonstrate that large pre-trained vision-language models can enhance the performance of deep metric learning methods.

\subsection{Qualitative Results}

\begin{figure}[t]
	\centering
	\includegraphics[width=0.8\linewidth]{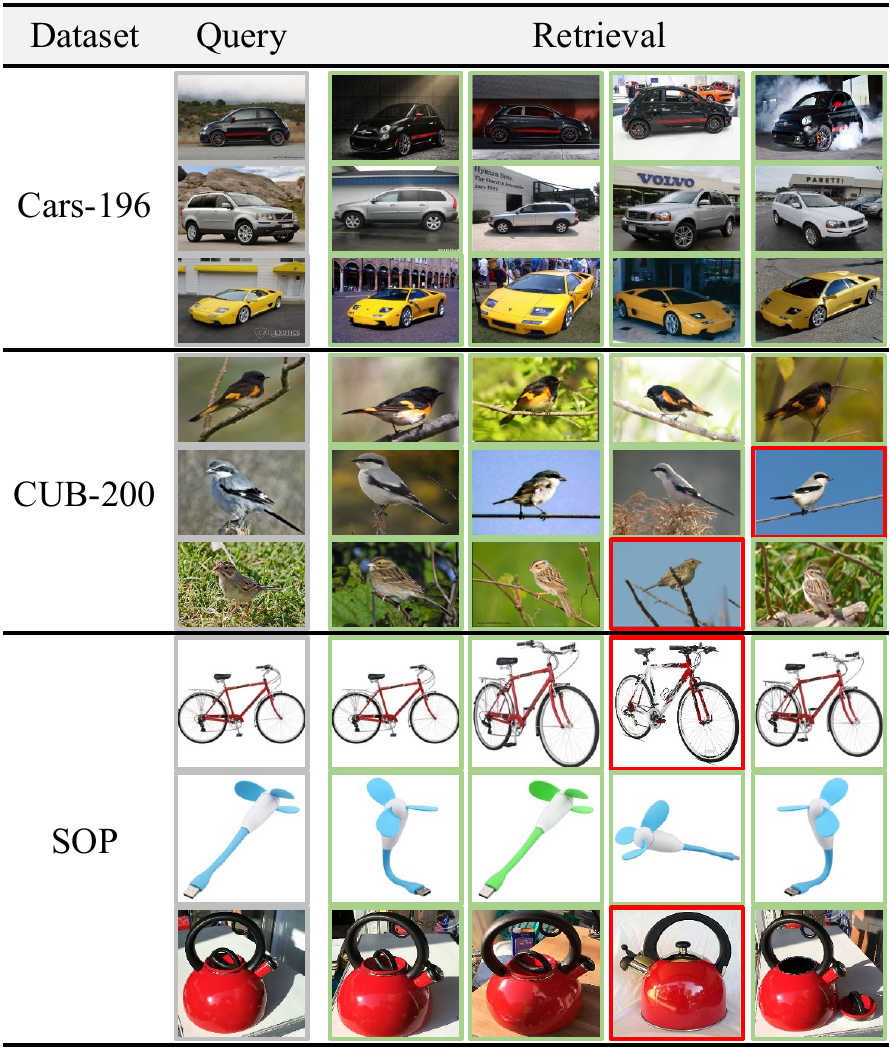}
	\caption{The qualitative retrieval results of the Anti-Collapse loss are demonstrated on the datasets CUB200, Cars196, and SOP (Stanford Online Products). The images enclosed in red-bordered boxes represent samples with unsuccessful retrievals. It can be observed these unsuccessful retrieval samples share similar characteristics with the successful retrieval samples.}
	\label{fig:qr}
\end{figure}

Enhancing image retrieval performance is the primary objective of optimizing the embedding space with the Anti-Collapse Loss. In addition to quantitative comparative experiments, we also present qualitative retrieval results of our method on three commonly used image retrieval datasets: CUB200, Cars196, and SOP (Stanford Online Products), as illustrated in Fig.~\ref{fig:qr}. These three datasets exhibit diverse sample poses, significant variations in sample quantities. Furthermore, there are unique image characteristics in each dataset. In CUB200, images have complex background with minor inter-class differences. In Cars196, there are noticeable color variations within sample classes. SOP features limited data per sample and substantial viewpoint changes. The qualitative results illustrated in Fig.~\ref{fig:qr} provide compelling evidence that our Anti-Collapse Loss consistently delivers outstanding retrieval performance, even across datasets exhibiting distinct characteristics.

\section{Discussion}
\label{sec:discussion}
The Anti-Collapse Loss proposed in this paper introduces a promising new direction for research in deep metric learning. It enhances the sparsity of feature clusters by maximizing the average coding rate of proxies, thereby alleviating the collapse issue in the embedding space. The method enhances inter-class differences by optimizing proxies, thereby improving classification and clustering performance. However, the relationship between proxies and samples of the same class is not the main focus of this paper, which may limit further improvement in method performance. Proxy-based methods optimize models based on the non-bijective similarity between sample-proxy pairs. This characteristic enhances computational efficiency but may also induce the isotropic distribution of features among samples of the same class due to the guiding role of proxies. The discriminative ability of intra-class samples significantly impacts the performance of classification and clustering methods. In future work, we will focus on addressing the challenges faced by proxy-based methods in handling the local relationships among samples of the same class, aiming to provide more comprehensive optimization for proxy-based methods.

\section{Conclusion}
\label{sec:conclusion}
In this work, we designed a novel loss function called Anti-Collapse Loss to address the issue of embedding space collapse in existing deep metric learning methods, which was caused by insufficient attention to the global structure of the embedding space. Specifically, we developed three versions of this loss, two of which were aimed at class proxies, and one was targeted at samples. These loss functions optimized the spatial structure of all samples or class proxies in the embedding space by maximizing their coding rate. Our proposed approach performed remarkably well on three commonly used image retrieval datasets, further highlighting the significance of optimizing the overall structure of the embedding space.

\section*{Acknowledgments}

This work was supported by the 173 Basic Strengthening Plan Technology Field Fund Project (2022-JCJQ-JJ-0221) and National Defense Science and Technology Commission Basic Research Project (JCKY2021208B043).

\bibliographystyle{IEEEtran}
\bibliography{main}

\end{document}